\documentclass{article}

\usepackage{tcolorbox}
\usepackage{listings}

\tcbuselibrary{listings, breakable}
\usepackage{enumitem}

\usepackage{makecell}   

\usepackage{xcolor}
\usepackage{tabularx, booktabs}

\usepackage{arxiv}

\usepackage{hyperref}       
\usepackage{url}            
\usepackage{amsfonts}       
\usepackage{nicefrac}       
\usepackage{microtype}      
\usepackage{lipsum}		
\usepackage{graphicx}
\usepackage{natbib}
\usepackage{doi}
\usepackage{amsmath}
\usepackage{caption}
\usepackage{threeparttable}
\usepackage{multirow}

\title{ 
\raisebox{-0.45\height}{\includegraphics[height=7.5 em]{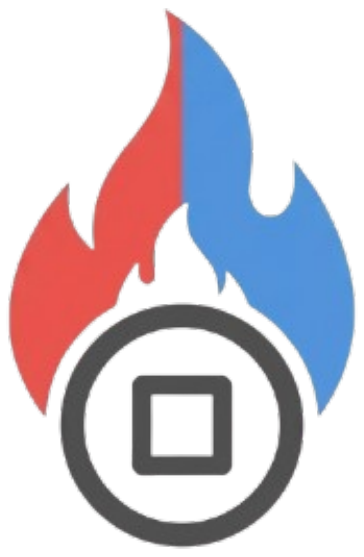}}%
  \hspace{-20 pt}%
\textbf{FIRE}: A Comprehensive Benchmark for \underline{\textbf{F}}inancial \underline{\textbf{I}}ntelligence and \underline{\textbf{R}}easoning \underline{\textbf{E}}valuation.
}

\date{} 					

\author{ 
    Xiyuan Zhang\textsuperscript{1*}, Huihang Wu\textsuperscript{2*}, Jiayu Guo\textsuperscript{3*} ,
     Zhenlin Zhang\textsuperscript{2}, Yiwei Zhang\textsuperscript{2}\\[0.05cm]\textbf{Liangyu Huo\textsuperscript{1},}
 \textbf{Xiaoxiao Ma\textsuperscript{1},} \textbf{Jiansong Wan\textsuperscript{1}, }
    \textbf{Xuewei Jiao\textsuperscript{1}, }\textbf{Yi Jing\textsuperscript{1}, }\textbf{Jian Xie\textsuperscript{1}} \\[0.15cm]
    \textsuperscript{1}Du Xiaoman Technology \\
    \textsuperscript{2}The PBC School of Finance, Tsinghua University \\
    \textsuperscript{3}The School of Finance, Renmin University of China \\
    \textsuperscript{*}Co-First Authors
}

\renewcommand{\headeright}{}

\hypersetup{
pdftitle={A template for the arxiv style},
pdfsubject={q-bio.NC, q-bio.QM},
pdfauthor={David S.~Hippocampus, Elias D.~Striatum},
pdfkeywords={First keyword, Second keyword, More},
}

\begin{document}
\maketitle
 
\begin{abstract}

The rapid advancement of Large Language Models (LLMs) has generated increasing interest in their potential applications within the financial domain. Despite this progress, rigorously and systematically evaluating the evolving LLMs’ ability to address complex and realistic financial tasks remains an underexplored challenge. Although several benchmarks have been proposed to assess the financial capabilities of LLMs, existing benchmarks often suffer from coarse or insufficient categorization, leading to substantial overlap across tasks while neglecting critical financial domains. Moreover, most of them either emphasize superficial understanding of financial terminology or adapt traditional Natural Language Processing (NLP) tasks to financial contexts. As a result, they fall short of evaluating models’ capacity to perform realistic financial scenario tasks. These limitations hinder an accurate assessment of LLMs’ true financial proficiency. To address these gaps, we introduce FIRE, a comprehensive benchmark designed to evaluate both the theoretical financial knowledge of LLMs and their ability to handle practical business scenarios. For theoretical assessment, we curate a diverse set of examination questions drawn from widely recognized financial qualification exams, enabling evaluation of LLMs’ deep understanding and application of financial knowledge. In addition, to assess the practical value of LLMs in real-world financial tasks, we propose a systematic evaluation matrix that categorizes complex financial domains and ensures coverage of essential subdomains and business activities. Based on this evaluation matrix, we collect 3,000 financial scenario questions, consisting of closed-form decision questions with reference answers and open-ended questions evaluated by predefined rubrics.
We conduct comprehensive evaluations of state-of-the-art LLMs on the FIRE benchmark, including XuanYuan 4.0, our latest financial-domain model, as a strong in-domain baseline.
These results enable a systematic analysis of the capability boundaries of current LLMs in financial applications. We publicly release the benchmark questions and evaluation code to facilitate future research.

\textbf{GitHub:} \url{https://github.com/FIRE-Bench/FIRE-Bench}

\end{abstract}    

\vspace{-0.5mm}
\section{Introduction}
Large Language Models (LLMs) have made remarkable progress since the initial release of ChatGPT \citep{openai2022chatgpt}. Their capabilities have rapidly expanded from basic Natural Language Processing (NLP) and conversational response generation to advanced mathematical reasoning \citep{hurst2024gpt, guo2025deepseek, yang2025qwen3}, multimodal reasoning \citep{comanici2025gemini}, and function calling \citep{team2025longcat, feng2025retool}. More recently, LLMs have demonstrated strong agentic capabilities \citep{team2025kimi}, enabling direct multi-turn interaction with real-world systems through access to diverse APIs and external tools. These advances have substantially accelerated the adoption of LLMs in the financial domain, motivating both researchers and practitioners to explore their potential to transform financial services.

Beyond general-purpose models, a growing body of financial domain–specific LLMs has emerged \citep{wu2023bloomberggpt, zhang2023xuanyuan, zhu2025dianjin, zheng2025agentar, hu2025finsearchcomp}. Through continued pretraining or fine-tuning on finance-specific corpora, these models are designed to address specialized financial tasks. At the same time, the enhanced reasoning and tool-use capabilities of modern LLMs further increase their suitability for complex financial problems that require deep document understanding, data analysis, and multi-step decision-making \citep{liu2025findabench}. As a result, LLM applications in finance now span the generation of automated investment research reports, end-to-end intelligent customer service, assisted design of sophisticated financial products, and large-scale risk and compliance monitoring. Collectively, these developments highlight the disruptive potential of LLMs and signal a fundamental transformation of the financial industry \citep{li2025investorbench, yu2024fincon, wang2025financial}.

Alongside these advances, the systematic and robust evaluation of LLMs’ financial capabilities has become increasingly critical, underscoring the need for well-designed and representative benchmarks. In response, numerous financial benchmarks have been proposed to assess model performance on finance-specific tasks \citep{zhu2024benchmarking, FinanceIQ, zheng2025agentar, zhang2023fineval, xie2023pixiu, xie2024finben, liu2025findabench}, including financial terminology explanation, information extraction, and financial text generation.

However, existing benchmarks exhibit several notable limitations. First, the majority of benchmarks focus on conventional NLP tasks that are only superficially adapted to financial contexts, and therefore fail to evaluate a models’ abilities to solve authentic financial business problems. Such evaluations capture only a limited subset of real-world financial practice. In reality, financial operations form a complex value chain involving multiple stages, such as insight generation, decision-making, and risk control. Effective models must not only possess financial knowledge, but also be able to operationalize it—integrating domain expertise into business workflows and solving practical problems in an end-to-end manner. Second, many existing benchmarks lack fine-grained and principled categorization of evaluation scenarios, leading to substantial task overlap while leaving important financial domains underrepresented. More fundamentally, benchmark results are often weakly connected to business value, which is the central concern of financial institutions. High benchmark scores offer limited insight into whether a model can improve credit approval efficiency, enhance marketing conversion rates, or detect emerging fraud patterns. This misalignment between evaluation metrics and real-world business objectives makes the Return On Investment (ROI) of AI deployment difficult to quantify, thereby impeding large-scale adoption of LLMs in core financial applications.

To address these challenges, we introduce FIRE, a comprehensive benchmark for Financial Intelligence and Reasoning Evaluation. FIRE consists of two complementary categories of evaluation tasks. The first category focuses on financial knowledge assessment, leveraging up-to-date certification examination questions from major professional qualification exams to evaluate LLMs’ depth of understanding and their ability to apply core financial concepts. The second category targets real-world financial operations and comprises thousands of scenario-based problems collected from practical financial business activities, designed to rigorously assess LLMs’ practical and operational problem-solving capabilities.

To enable a comprehensive and business-oriented evaluation, we further propose a novel matrix-style evaluation framework that systematically integrates vertical business dimensions with horizontal capability dimensions. This framework guides both the construction and evaluation of 3,000 real-world financial business problems. Of these, 1,000 questions are paired with high-quality reference answers, enabling objective and reliable automatic evaluation. For the remaining 2,000 open-ended questions without reference answers, we develop detailed, question-specific rubrics and introduce an automated evaluation pipeline to support scalable and consistent assessment. Collectively, FIRE provides a structured, rigorous, and application-driven framework for benchmarking the real-world financial intelligence and reasoning capabilities of LLMs.

Using FIRE, we conduct extensive evaluations of state-of-the-art general-purpose and finance-specialized LLMs, including our newly developed flagship financial model, XuanYuan 4.0. The results reveal critical insights into the strengths, limitations, and capability gaps of current LLMs in complex financial domains.

The remainder of this paper is organized as follows. Section 2 presents the design principles and detailed components of the FIRE benchmark. Section 3 introduces the XuanYuan 4.0 model, our newly developed flagship financial-domain model, as a competitive baseline. Section 4 reports experimental results and comparative analyses across representative models evaluated on FIRE. Section 5 concludes the paper.
\vspace{-0.5mm}
\section{FIRE Benchmark}
\label{method}

\begin{figure*}[h]
\centerline{\includegraphics[scale=0.5]{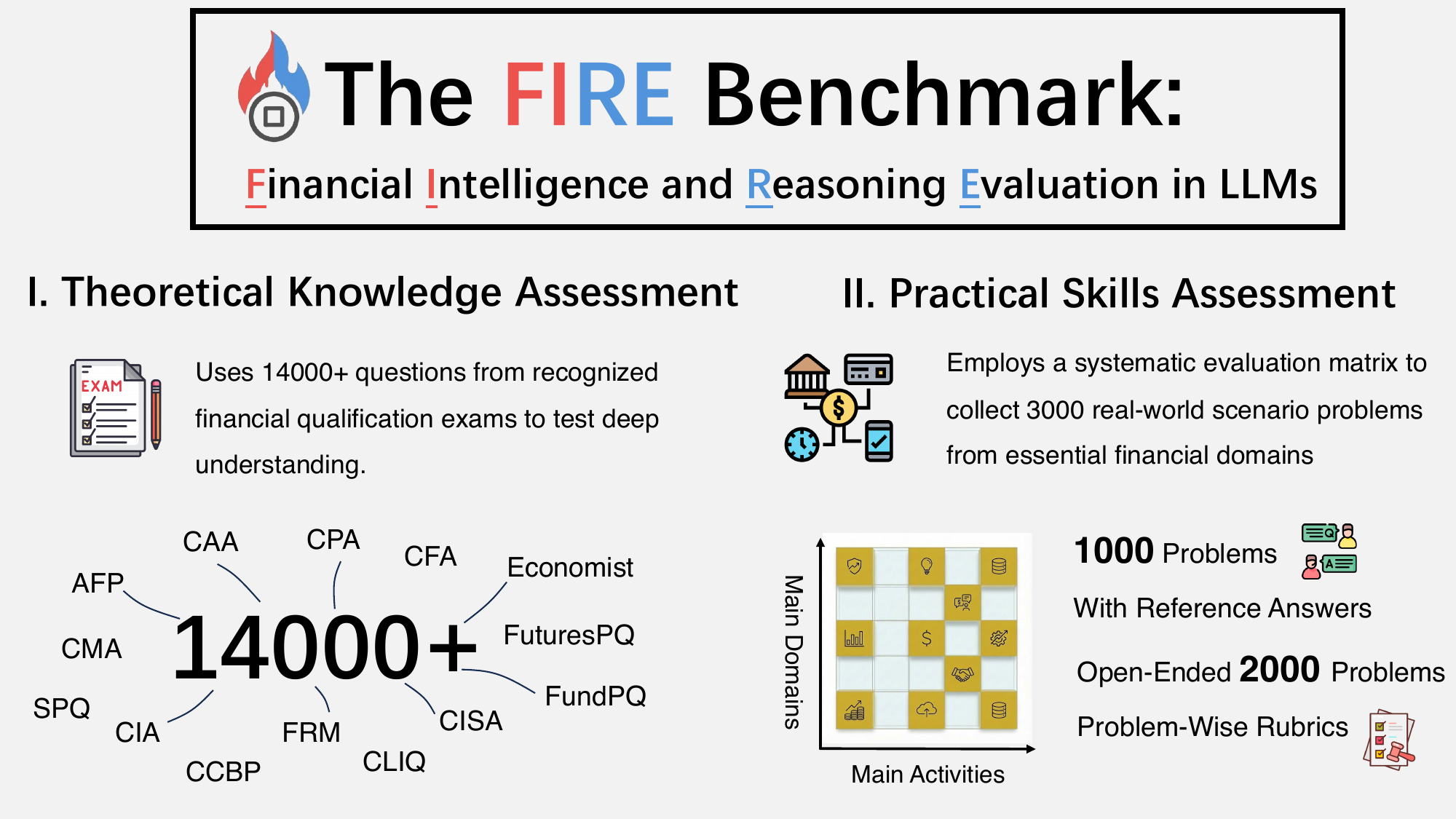}}
\caption{Overview of FIRE Benchmark}
\label{fig1}
\end{figure*}

In this section, we present a comprehensive description of FIRE, which is jointly developed by Du Xiaoman Technology—a leading company in China’s Internet finance industry—the PBC School of Finance (PBCSF) at Tsinghua University, and the School of Finance at Renmin University of China. This close industry–academia collaboration ensures that FIRE incorporates professional insights from both financial practitioners and academic experts, enabling rigorous evaluation of large language models with respect to both theoretical financial knowledge and practical, scenario-driven problem-solving capabilities in real-world financial environments.

As illustrated in Figure \ref{fig1}, FIRE comprises two main categories of problems. The first category, financial knowledge assessment, consists of questions collected from recognized professional financial qualification examinations and is designed to evaluate models’ deep understanding and application of core financial concepts. The second category, practical skills assessment, includes problems derived from real-world business operations across essential financial domains, aiming to assess models’ ability to solve realistic, operationally grounded financial problems.

In the following subsections, we provide a detailed taxonomy of all problem categories and describe the corresponding evaluation protocols and criteria.

\begin{table}[ht]
\centering
\small 
\setlength{\tabcolsep}{6pt} 
\renewcommand{\arraystretch}{1.4}
\begin{tabularx}{\textwidth}{l X}
\toprule
 \textbf{Exam} & \textbf{Professional Scope \& Description} \\
\midrule
\multicolumn{2}{l}{\textit{International Professional Certifications}} \\
CFA & Global certification in investment management and financial analysis.  \\
CIA & International standard for professional internal auditing.  \\
CISA & Global benchmark for information systems auditing, control, and security.  \\
CMA & Worldwide certification in management accounting and financial strategy.  \\
FRM & International specialization in financial risk management and analysis.  \\
\midrule
\multicolumn{2}{l}{\textit{Domestic (China) Professional Certifications}} \\
AFP & Associate Financial Planner certification for practitioners in China.  \\
CAA & National qualification for certified actuaries in China.  \\
CCBP & Official national certification for China's banking sector practitioners.  \\
CLIQ & Professional qualification for the Chinese life insurance industry.  \\
CPA & National professional certification for Certified Public Accountants in China.  \\
Economist & National proficiency exam (Junior, Intermediate, and Senior levels).  \\
FundPQ & Mandatory qualification for fund industry practitioners in China.  \\
FuturesPQ & Entry-level qualification for the Chinese futures market.  \\
SPQ & Official qualification for practitioners in the Chinese securities industry.  \\
\bottomrule
\end{tabularx}
\caption{Classification and Description of Professional Qualification Examination Sources in FIRE.}
\label{tab:fire_exam_sources}
\end{table}

\subsection{Financial Qualification Problems}
The financial theoretical knowledge assessment set comprises approximately 14,000 carefully curated questions drawn from major professional qualification certifications in finance, enabling a comprehensive and well-balanced evaluation of financial expertise. The benchmark spans 14 core financial examinations, including AFP (Associate Financial Planner), CAA (China Association of Actuaries), CFA (Chartered Financial Analyst), CIA (Certified Internal Auditor), CISA (Certified Information Systems Auditor), CMA (Certified Management Accountant), CPA (Certified Public Accountant), FRM (Financial Risk Manager), CLIQ (China Life Insurance Qualification), FundPQ (Fund Practitioner Qualification), FuturesPQ (Futures Practitioner Qualification), Economist, SPQ (Securities Practitioner Qualification), and CCBP (Certification of China Banking Professional). Detailed descriptions of each examination are provided in Table \ref{tab:fire_exam_sources}. Across all exam types, the collection maintains a dedicated focus on financial industry security and regulatory compliance.

For domestic certifications, the benchmark provides extensive coverage of banking, securities, fund, and futures practitioner qualifications, as well as the Economist, insurance practitioner, Chinese actuary, and CPA certifications. It places particular emphasis on China-specific regulatory frameworks and compliance practices, including the Securities Law, anti–money laundering regulations, and credit governance requirements. Representative examples include the “Laws, Regulations, and Comprehensive Professional Capabilities” module in banking qualifications and the “Basic Laws and Regulations of the Securities Market” component in securities examinations.

Regarding international certifications, the benchmark systematically incorporates CFA (Ethical and Professional Standards), FRM (Risk and Investment Management), CIA (Internal Audit Practice), and CISA (Information Systems Auditing). These components emphasize adherence to globally recognized ethical principles and international risk management frameworks, such as the CFA standards governing conflicts of interest, fiduciary responsibilities, and duties to clients.

\begin{table*}[ht]
\centering
\scriptsize
\renewcommand{\arraystretch}{1.5} 
\setlength{\tabcolsep}{4pt}

\begin{tabularx}{\textwidth}{>{\bfseries}l >{\raggedright\arraybackslash}p{2.5cm} X X X X}
\toprule
 \textbf{Sector} & \textbf{Business Line} & \textbf{Insight \& Decision} & \textbf{Product \& Marketing} & \textbf{Service \& Operation} & \textbf{Risk \& Compliance} \\
\midrule

\multirow{3}{*}{Banking} 
& Corporate Finance & Financial analysis, Industry research, Credit rating & Credit product design, Supply chain solutions & Corporate consulting, Process guidance & Credit risk, Anti-fraud, Policy compliance \\ \cline{2-6}
& Retail Finance & User profiling, Behavior analysis & Loan \& card design, Pricing, Marketing copy & Intelligent CS, Process guidance, Complaints & Personal credit risk, Anti-fraud, Suitability \\ \cline{2-6}
& Markets \& Treasury & Macro analysis, Trend forecasting & Structured products, Trading strategies & Execution support, Fund clearing, Reports & Market \& Liquidity risk, Trading compliance \\
\midrule

\multirow{3}{*}{Insurance} 
& Property & Risk factor analysis, Premium research & Product design, Policy drafting, Renewal & Online claims, Consultation, Loss assessment & Underwriting risk, Claims anti-fraud, Liability \\ \cline{2-6}
& Life & Health analysis, Life table research & Product design, Pricing, Agent empowerment & Intelligent underwriting, Policy services & Underwriting risk, Claims anti-fraud, Compliance \\ \cline{2-6}
& Reinsurance & Risk exposure, Catastrophe modeling & Structure design, Pricing support & Statement processing, Reconciliation & Risk assessment, Solvency, Clause review \\
\midrule

\multirow{2}{*}{Securities} 
& Invest. Banking & Industry research, M\&A screening & Financing solutions, Transaction structure & Document drafting, Working papers & Due diligence, Risk ID, Verification \\ \cline{2-6}
& Proprietary & Market/Stock research, Backtesting & Quant strategies, R\&D, Data services & Code \& log generation, Institutional Q\&A & Trading risk, Strategy failure, Compliance \\
\midrule

\multirow{2}{*}{Fund} 
& Private Equity & BP screening, Sector scanning, Due Diligence & Fund structure, Investment strategy, PPM & Post-investment reports, Disclosure & Project risk, Valuation, Agreement review \\ \cline{2-6}
& Distressed Assets & Asset valuation, Opportunity analysis & Disposal plan design, Restructuring & Due diligence summary, Process support & Disposal risk, Legal risk identification \\
\midrule

\multirow{2}{*}{Futures} 
& Brokerage & Market interpretation, Basis analysis & -- & Trading rules, Settlement, Software guide & Margin risk, Liquidation warning, Suitability \\ \cline{2-6}
& Risk Mgmt. & Basis trading, Corporate exposure & Hedging \& OTC derivatives design & Strategy consulting, Evaluation reports & OTC pricing, Hedging risk, Compliance \\
\midrule

Trust & Invest. \& Fin. & Project due diligence, Credit analysis & Trust structure, Transaction innovation & Post-establishment mgmt, Disclosure & Project risk, Repayment forecast, Contract review \\
\midrule

\multirow{2}{*}{FinTech} 
& Online Payments & Transaction behavior, User profiling & Innovative products, Scenario solutions & Intelligent CS, Dispute resolution, Clearing & AML risk, Anti-fraud, Payment compliance \\ \cline{2-6}
& Tech Solutions & Client demand, Competitor analysis & Technical design, White paper writing & Technical support, Design documents & Technical risk, Security \& Data privacy \\
\midrule

\multirow{2}{*}{General} 
& Asset Mgmt. & Invest. research, Asset allocation & Wealth products, Portfolio design & Post-investment, Performance attribution & Portfolio risk, Disclosure, Operational \\ \cline{2-6}
& Wealth Mgmt. & Allocation advice, Trend interpretation & Robo-advisory, Family wealth planning & Dedicated advisory, Meeting minutes & Client risk, Suitability, AML, Risk alerts \\
\bottomrule
\end{tabularx}
\caption{Financial Business Evaluation Matrix: Mapping Sectors to Core Competencies.}
\label{tab:finance_matrix}
\end{table*}

\subsection{Real-World Financial Scenario Problems}
Finance is an inherently multifaceted domain, characterized by specialized subfields with divergent regulatory and operational mandates. To systematically evaluate the proficiency of LLMs in navigating these complexities, we propose a two-dimensional Financial Application Scenario Evaluation Matrix as detailed in Table \ref{tab:finance_matrix}.

The row dimension of this matrix delineates eight primary financial sectors—Banking, Insurance, Securities, Fund Management, Futures, Trust, FinTech, and General Finance—which are further granularized into seventeen business subcategories. The column dimension captures four fundamental functional pillars: Insight \& Decision, Product Design \& Marketing, Service \& Operation, and Risk \& Compliance. These axes represent the functional backbone of the industry and collectively dictate the efficacy of value creation.

This framework provides a rigorous decomposition of the tasks AI systems must master within specific scenarios. Beyond its utility as a benchmark, this matrix serves as a principled roadmap for the strategic curation of training data and the targeted enhancement of financial LLMs. Furthermore, it offers a blueprint for financial institutions to orchestrate end-to-end digital transformations by defining clear trajectories for scenario deployment.

In alignment with this framework, we curated a dataset of 3,000 high-fidelity, scenario-based questions designed to populate the matrix cells. This methodology ensures a granular, multi-dimensional assessment of LLM performance across both vertical business domains and horizontal functional competencies.

\subsection{Evaluation Criteria}
Evaluation of the financial qualification dataset is conducted using a binary scoring objective. As all questions in this set are multiple-choice with verified ground-truth labels, model performance is measured by the direct alignment of the predicted option with the gold-standard answer. A correct prediction is assigned a score of 1, whereas any incorrect or non-matching response receives a score of 0.

For real-world financial scenario tasks, the evaluation methodology diverges based on the availability of reference answers:

\begin{itemize}
    \item \textbf{Tasks with Reference Answers:} For scenarios with objective ground truth, models are prompted to generate outputs in a structured JSON format. We utilize an automated extraction pipeline to isolate the final response and perform an exact-match verification against the ground truth. Consistent with the qualification set, successful matches are awarded a score of 1, and failures are scored as 0.
    \item \textbf{Open-Ended Tasks:} For tasks lacking ground truths, we employ a rubric-based automated grading approach. Detailed, task-specific evaluation rubrics were constructed to capture nuanced dimensions of financial expertise. A dedicated scoring model, trained specifically to interpret these rubrics, analyzes the problem description and model output to produce a quantitative quality score. Comprehensive details regarding the architectural training of the scoring model and the formulation of the evaluation rubrics are documented in Appendix A.
\end{itemize}


\subsection{Scoring Model for Open-Ended  Questions}

Evaluating open-ended financial tasks without gold-standard references typically relies on an LLM-as-a-judge paradigm. However, prior studies have shown that such evaluations are often unstable, being highly sensitive to response length, prompt phrasing, and evaluator-specific biases, which can lead to inconsistent scoring of identical outputs.

To address this issue, we adopt a structured, rubric-based scoring framework tailored to the financial domain. Instead of relying on holistic judgments, we decompose expert financial evaluation into explicit, fine-grained criteria with clearly defined scoring standards. This design improves evaluation stability and interpretability, and enables reliable assessment under domain-specific regulatory and risk constraints.

\begin{figure*}[h]
\centerline{\includegraphics[scale=0.5]{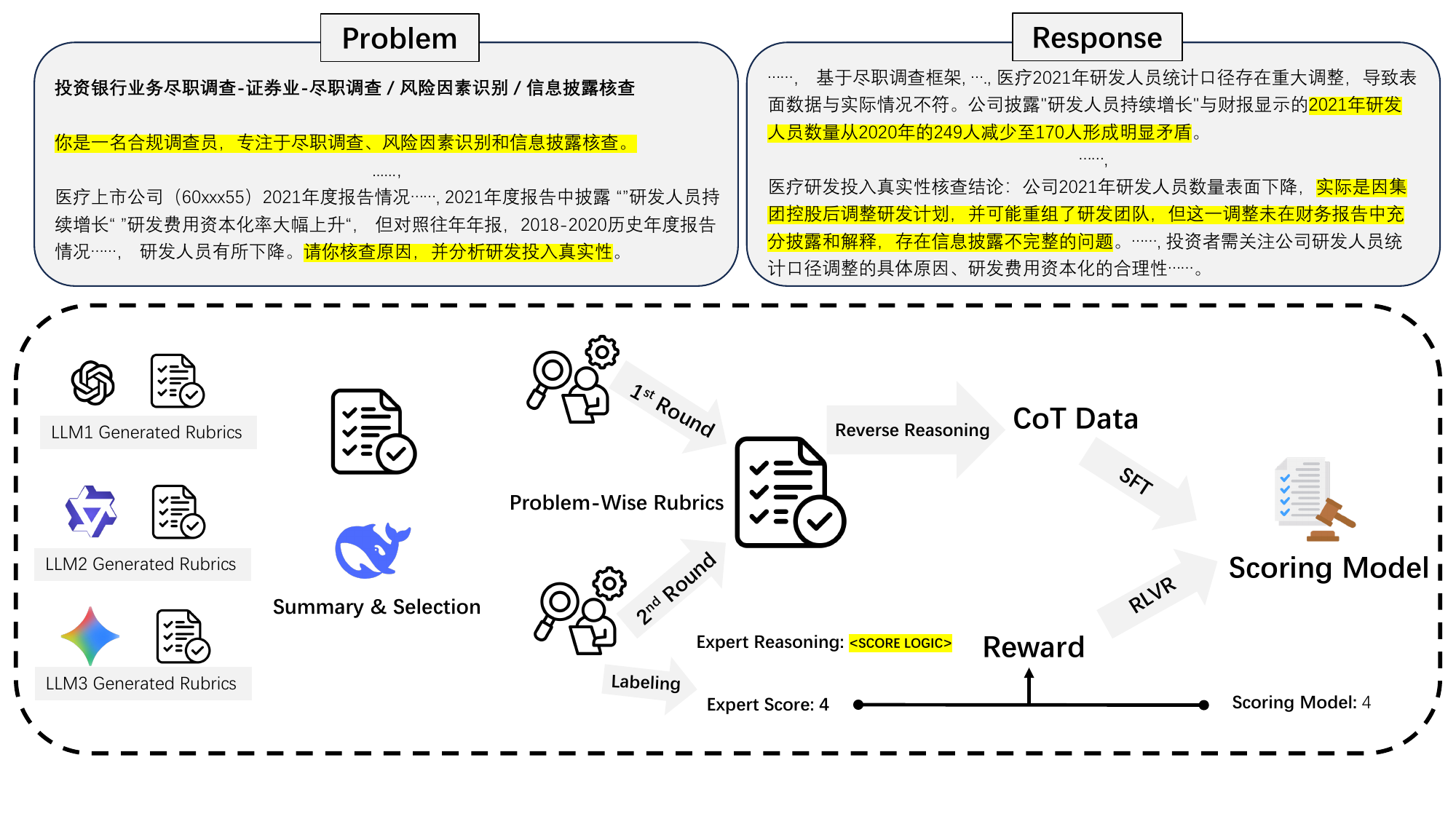}}
\caption{Problem-Wise Rubrics Generation and Training Pipeline of the Scoring Model}
\label{fig2}
\end{figure*}

Based on this framework, we further train a dedicated scoring model to perform automated, consistent evaluation. The overall training pipeline of the scoring model is illustrated in Figure \ref{fig2}, with detailed specifications provided in Appendix A.
\section{A Strong Baseline: XuanYuan 4.0}
\label{section-model}

To facilitate further research and provide a standardized performance reference, we develop XuanYuan 4.0, a large language model specifically enhanced for the financial domain alongside the FIRE benchmark. Developed by Du Xiaoman Technology, XuanYuan 4.0 is the latest flagship in the XuanYuan series—a 36B-parameter dense model initialized from the Seed-OSS-36B-Base \citep{seed2025seed-oss}. 

The development of XuanYuan 4.0 follows a rigorous multi-stage pipeline. Leveraging an extensive repository of high-quality financial corpora, we employ an additional CPT strategy that integrates curated domain data with an annealed training schedule. To preserve model stability, we incorporate a self-regularization objective based on the Kullback–Leibler (KL) divergence relative to a reference model. This approach enhances financial knowledge density and reasoning depth while mitigating model drift, thereby ensuring robust generalization and iterative scalability.

Following CPT, the model undergoes Supervised Fine-Tuning (SFT) using a high-fidelity dataset spanning mathematics, STEM, and agentic tasks to bolster its foundational reasoning. We further refine the model through Reinforcement Learning with Verifiable Rewards (RLVR), specifically targeting financial scenarios derived from internal business activities. This stage is optimized using the DAPO algorithm \citep{yu2025dapo} to align model outputs with complex financial logic.

Empirical evaluations demonstrate that XuanYuan 4.0 achieves performance parity with leading closed-source models across authoritative financial examination leaderboards. Concurrently, its general-purpose capabilities remain highly competitive with mainstream open-source models of a comparable parameter scale. In the subsequent section, we present a comprehensive analysis, evaluating the performance of several state-of-the-art models alongside XuanYuan 4.0 on the FIRE benchmark.
\section{Experiments}
\subsection{Evaluated Models and Setups}
We conducted an extensive evaluation of both leading proprietary and open-source LLMs to establish a comprehensive performance baseline on our benchmark. The models selected represent the current state-of-the-art models in the field: 
\begin{itemize} 
\item Proprietary Models: The cohort of closed-source models includes Gemini 3.0 Pro \citep{gemini3.0}, GPT 5.2 \citep{gpt51}, and Claude 4.5 Sonnet \citep{sonnet}, and  Qwen3 Max \citep{qwen3max}, representing the frontier of high-parameter, commercially available systems. 
\item Open-Source Models: The open-weight landscape is represented by a diverse set of architectures, including DeepSeek V3.2 \citep{liu2025deepseek}, Kimi K2 Thinking \citep{k2thinking}, GLM 4.6 \citep{glm46}, Qwen3-235B-A22B-Thinking-2507 \citep{yang2025qwen3}, GPT-OSS-120B \citep{agarwal2025gpt}, and Seed-OSS-36B-Instruct \citep{seed2025seed-oss}. 
\item Financial Models: Dianjin-R1 \citep{zhu2025dianjin} and XuanYuan-FinX1, two specialized open-source financial models recognized for their robust reasoning capabilities.
\end{itemize} 


To ensure consistency across evaluations, we standardized the inference hyperparameters whenever possible. Specifically, we adopted the officially recommended inference settings for each model when available; otherwise, we defaulted to a sampling temperature of 0.6 and a top-$p$ value of 0.8.

\begin{table*}[ht]
\centering
\scriptsize
\renewcommand{\arraystretch}{1.5}
\setlength{\tabcolsep}{1.2 pt}
\begin{tabularx}{\textwidth}{l c ccccc cccccc cc}
\toprule
\multirow{2}{*}{\textbf{Exam}} 
& \multirow{2}{*}{\textbf{XuanYuan}} 
& \multicolumn{5}{c}{\textbf{Proprietary Models}} 
& \multicolumn{6}{c}{\textbf{Open-Source Models}} 
& \multicolumn{2}{c}{\textbf{Financial Models}} \\
\cmidrule(lr){3-7} \cmidrule(lr){8-13} \cmidrule(lr){14-15}

& 4.0
& \textbf{Gemini} 
& \textbf{GPT} 
& \textbf{Claude} 
& \textbf{Qwen3} 
& \textbf{Doubao} 
& \textbf{DeepSeek} 
& \textbf{Kimi} 
& \textbf{GLM} 
& \textbf{Qwen3} 
& \textbf{GPT} 
& \textbf{Seed}

& \textbf{Dianjin}
& \textbf{XuanYuan}\\

&
& 3.0 
& 5.2 
& 4.5 
& Max 
& Seed-1.6 

& V3.2 
& K2-Think 
& 4.6 
& 235B 
& OSS-120B 
& OSS-36B 

& R1
&FinX1\\
\midrule

China Life Insurance              & 89.95 & 89.40 & 83.95 & 81.38 & 85.87 & 87.65 & 84.67 & 82.10 & 81.38 & 84.57 & 70.78 & 84.67 & 84.87 & 68.51 \\
China Actuary (CAA)               & 93.17 & 91.16 & 94.88 & 69.77 & 65.89 & 93.95 & 95.35 & 94.88 & 89.77 & 94.42 & 91.16 & 88.83 & 81.86 & 30.23 \\
Fund Practitioner                 & 95.51 & 95.78 & 87.51 & 85.18 & 92.66 & 95.50 & 91.19 & 90.17 & 86.70 & 91.82 & 75.47 & 92.18 & 90.29 & 70.44 \\
Futures Practitioner              & 90.46 & 89.83 & 79.86 & 79.17 & 86.15 & 91.90 & 84.70 & 84.30 & 82.13 & 85.19 & 70.58 & 87.16 & 84.10 & 54.68 \\
Securities (SPQ)                  & 94.72 & 92.47 & 82.33 & 84.41 & 92.22 & 93.45 & 90.18 & 88.77 & 84.41 & 89.31 & 70.77 & 89.42 & 89.53 & 64.55 \\
Banking (CCBP)                    & 91.61 & 90.06 & 82.19 & 79.21 & 89.04 & 92.16 & 85.35 & 85.67 & 81.28 & 86.92 & 68.23 & 88.05 & 86.54 & 64.03 \\
Economist (Jr/Int/Sr)             & 90.47 & 89.72 & 83.40 & 78.26 & 87.44 & 90.64 & 85.11 & 86.17 & 79.84 & 85.24 & 66.53 & 87.87 & 85.11 & 61.92 \\
Financial Planner (AFP)           & 92.36 & 92.29 & 86.36 & 80.24 & 79.84 & 92.29 & 88.74 & 88.74 & 83.79 & 86.96 & 77.47 & 86.75 & 83.20 & 60.86 \\
CFA                               & 92.48 & 95.03 & 91.65 & 85.61 & 83.90 & 92.00 & 88.28 & 91.30 & 87.21 & 90.59 & 85.79 & 87.56 & 79.57 & 63.23 \\
Internal Auditor (CIA)            & 91.57 & 91.25 & 79.84 & 79.09 & 82.83 & 86.12 & 77.57 & 80.80 & 74.14 & 84.41 & 69.96 & 81.55 & 73.38 & 65.58 \\
IS Auditor (CISA)                 & 90.85 & 91.29 & 82.81 & 82.37 & 84.00 & 83.48 & 74.33 & 81.03 & 74.55 & 86.16 & 70.98 & 77.45 & 72.99 & 69.86 \\
Management Accountant (CMA)       & 95.49 & 96.76 & 89.90 & 83.05 & 83.43 & 92.00 & 90.10 & 91.62 & 89.33 & 92.57 & 84.19 & 90.47 & 84.00 & 66.47 \\
Public Accountant (CPA)           & 91.87 & 90.77 & 78.24 & 72.07 & 82.29 & 93.12 & 85.69 & 83.40 & 78.05 & 85.78 & 54.86 & 88.39 & 82.72 & 49.55 \\
Risk Manager (FRM)                & 92.57 & 92.29 & 91.65 & 82.01 & 82.80 & 90.57 & 88.22 & 91.43 & 88.87 & 90.79 & 85.86 & 83.51 & 79.22 & 48.60 \\

\midrule
\textbf{Overall Average}          & 92.15 & 91.43 & 83.00 & 79.01 & 85.88 & 91.78 & 86.10 & 85.95 & 81.70 & 87.31 & 68.94 & 87.55 & 84.13 & 60.10 \\
\bottomrule
\end{tabularx}
\caption{Performance Comparison of Proprietary and Open-Source Models on Financial Certification Examinations (\%).}
\label{tab:translated_exam_results}
\end{table*}

\begin{table*}[ht]
\centering
\scriptsize
\renewcommand{\arraystretch}{1.5} 
\setlength{\tabcolsep}{2.5 pt} 
\begin{tabularx}{\textwidth}{l c ccccc cccccc cc}
\toprule
\multirow{2}{*}{\textbf{Sector}} 
& \multirow{2}{*}{\textbf{XuanYuan}} 
& \multicolumn{5}{c}{\textbf{Proprietary Models}} 
& \multicolumn{6}{c}{\textbf{Open-Source Models}} 
& \multicolumn{2}{c}{\textbf{Financial Models}} \\
\cmidrule(lr){3-7} \cmidrule(lr){8-13} \cmidrule(lr){14-15}

& 4.0
& \textbf{Gemini} 
& \textbf{GPT} 
& \textbf{Claude} 
& \textbf{Qwen3} 
& \textbf{Doubao} 
& \textbf{DeepSeek} 
& \textbf{Kimi} 
& \textbf{GLM} 
& \textbf{Qwen3} 
& \textbf{GPT} 
& \textbf{Seed}

& \textbf{Dianjin}
& \textbf{XuanYuan}\\

&
& 3.0 
& 5.2 
& 4.5 
& Max 
& Seed-1.6 

& V3.2 
& K2-Think 
& 4.6 
& 235B 
& OSS-120B 
& OSS-36B 

& R1
&FinX1\\
\midrule

Banking           & 78.55 & 73.58 & 77.41 & 71.33 & 71.43 & 75.49 & 74.87 & 72.57 & 72.10 & 72.39 & 70.12 & 74.78 & 62.08 & 59.53 \\
Insurance         & 80.72 & 79.82 & 82.53 & 73.42 & 76.91 & 79.64 & 77.86 & 74.85 & 77.11 & 76.45 & 76.81 & 82.79 & 63.64 & 73.95 \\
Securities        & 80.29 & 80.62 & 84.61 & 77.29 & 83.58 & 81.27 & 81.51 & 80.78 & 79.15 & 81.10 & 78.35 & 81.69 & 73.43 & 68.40 \\
Fund Management   & 85.21 & 88.73 & 91.20 & 85.36 & 86.23 & 82.74 & 86.27 & 86.97 & 83.80 & 85.21 & 83.45 & 87.29 & 75.35 & 75.00 \\
Futures           & 74.63 & 72.79 & 76.14 & 70.00 & 70.71 & 73.30 & 72.57 & 72.22 & 73.33 & 71.64 & 71.62 & 72.67 & 62.41 & 62.41 \\
Trust             & 68.38 & 72.06 & 75.00 & 50.74 & 71.21 & 65.44 & 63.97 & 63.97 & 66.18 & 66.17 & 57.35 & 70.00 & 68.94 & 58.09 \\
FinTech           & 73.17 & 65.37 & 60.70 & 47.94 & 60.06 & 69.80 & 64.19 & 64.72 & 63.53 & 62.34 & 61.70 & 64.63 & 56.44 & 62.23 \\
General Finance   & 85.79 & 83.80 & 80.36 & 77.73 & 80.35 & 85.71 & 82.38 & 81.75 & 81.60 & 84.65 & 77.03 & 82.29 & 77.96 & 73.84 \\
\midrule
\textbf{Overall Average} 
                  & 79.08 & 75.72 & 77.46 & 70.22 & 73.52 & 77.15 & 75.65 & 74.19 & 73.91 & 74.35 & 71.95 & 76.34 & 65.24 & 64.90 \\
\bottomrule
\end{tabularx}
\caption{Sector-Level Performance Comparison of Proprietary and Open-Source Models on Real-World Financial Business Scenarios(\%).}
\label{tab:sector_evaluation}
\end{table*}

\subsection{Main Results}

Table \ref{tab:translated_exam_results} summarizes the performance of the evaluated models. Despite the recognized rigor and difficulty of professional certification examinations, all models achieved relatively high scores, suggesting that state-of-the-art LLMs exhibit substantial capabilities in financial knowledge acquisition and domain-specific reasoning. Among the evaluated systems, Gemini 3.0 Pro demonstrated the strongest performance among general-purpose proprietary and open-source models.
Benefiting from high-quality financial corpora, XuanYuan 4.0 is introduced as a strong open-source baseline. It achieves the best performance among open-source models across the certification benchmarks, with overall results that are broadly comparable to Gemini 3.0 Pro.

Regarding the evaluation of real-world financial scenario problems, aggregated results over 3,000 cases are reported in Table \ref{tab:sector_evaluation}. Among top-tier proprietary models, GPT 5.2 achieves the strongest overall performance.
Notably, XuanYuan 4.0, after undergoing targeted financial instruction tuning, consistently outperforms its backbone model Seed-OSS-36B as well as other open-source baselines on financial scenario tasks. Despite its relatively modest scale of 36B parameters, XuanYuan 4.0 delivers overall performance that is highly comparable to GPT 5.2.
These results underscore the effectiveness of domain-specific alignment in enhancing practical financial reasoning and decision-making capabilities, while retaining significantly lower deployment and inference costs.

These findings indicate that current models perform exceptionally well on financial qualification exams, likely benefiting from standardized formats and accessible training data. However, this success does not translate to real-world financial scenarios, where proficiency remains limited. Such a disparity underscores a significant gap, necessitating further research to align theoretical mastery with practical operational requirements.
\section{Conclusions}
As LLMs increasingly permeate the financial sector, the necessity for rigorous, high-fidelity evaluation frameworks has become paramount. This paper introduces FIRE, a comprehensive benchmark designed to evaluate both the theoretical acumen and the practical problem-solving efficacy of LLMs within financial domains. By integrating professional certification standards with scenario-driven tasks modeled after real-world operations, FIRE bridges the divide between static academic metrics and the volatile requirements of industry deployment.

The evaluation results reveal a significant performance decoupling: while models often demonstrate mastery in knowledge-centric assessments, they frequently falter when confronted with complex, operational financial scenarios. This disparity highlights a fundamental limitation in current architectures—the inability to translate theoretical comprehension into reliable, actionable intelligence. These findings underscore the urgent need for domain-specific optimization, particularly in high-stakes environments where accuracy, robustness, and cost-efficiency are essential. Consequently, FIRE serves as a foundational framework for advancing the secure and scalable integration of LLMs within the global financial ecosystem.

\bibliographystyle{unsrtnat}
\bibliography{references}  

\appendix 

\clearpage


\section{Evaluation of Open-Ended Financial Scenario-Based Problems}




In the absence of gold-standard reference answers, the evaluation of LLMs on open-ended tasks has largely converged on the LLM-as-a-judge paradigm. Despite its widespread adoption, recent empirical studies have demonstrated that this approach suffers from substantial instability. Scoring outcomes are often highly sensitive to superficial factors such as response length, prompt formulation, and the inherent biases of the evaluator model, leading to inconsistent assessments even for semantically equivalent responses \citep{li2025generation}.

To mitigate these limitations, prior work has increasingly shifted toward rubric-based evaluation frameworks that explicitly structure the evaluation process. A representative example is OpenAI’s HealthBench, in which medical experts curated over 40,000 fine-grained evaluation criteria covering clinical correctness, safety, and regulatory compliance \citep{arora2025healthbench}. Such structured rubrics have been shown to significantly improve evaluation consistency, interpretability, and alignment with expert judgment.

Inspired by these advances, we design a case-specific, fine-grained scoring framework tailored to financial tasks. Financial decision-making is inherently high-stakes and subject to strict regulatory, risk-control, and compliance requirements, which generic evaluators often fail to adequately capture. We therefore decompose expert financial judgment into a set of explicit, observable evaluation dimensions, each associated with tiered scoring standards. This formulation enables transparent, reproducible, and reliable evaluation, which is essential for benchmarking LLMs in realistic financial scenarios.

Based on this rubric, we further train a dedicated scoring model to perform automated evaluation at scale. The training objectives and data construction process are described in detail in the following sections.

\subsection{Rubric Generation Pipeline} To derive stable, per-question scoring criteria for real-world financial benchmarks, we propose a multi-stage rubric generation pipeline. Unlike traditional methods that utilize a single generic rule for an entire task category, our approach accounts for the heterogeneity of financial tasks by generating a dedicated 1–5 point rubric for every individual question. This ensures granularity and interpretability during large-scale annotation. The pipeline consists of the following steps:

\begin{itemize} 
\item \textbf{Collaborative Multi-Model Generation:} We design task-specific prompts tailored to the problem type (e.g., risk assessment, compliance analysis). These prompts are optimized to produce structured criteria, including essential information requirements, logical reasoning chains, and relevant business constraints. By employing multiple LLMs collaboratively, we capture a diverse range of linguistic styles and error sensitivities, creating a rich candidate space for subsequent synthesis.
\item \textbf{Single-Model Synthesis:} A dedicated model synthesizes these candidates into a "draft rubric." This stage distills common elements and highlights distinctive key points, providing a high-quality baseline for human oversight.
\item \textbf{Expert Optimization:} Human review is the critical anchor of this pipeline. Domain experts refine the draft by externalizing "tacit knowledge"—correcting domain inaccuracies and logical gaps—thereby transforming subjective expertise into reproducible scoring rules.
\item \textbf{Empirical Validation via Model Sampling:} To ensure executability, we sample model-generated responses to the benchmark questions. Annotators then apply the draft rubrics to these samples. We specifically monitor for "unscorable" cases, ambiguities, or high inter-annotator variance. If discrepancies arise, the rubric is iteratively refined until the criteria are proven to be practically actionable.
\end{itemize}

This pipeline constructs a case-level system aligned with business logic, avoiding the ambiguity of template-based rubrics while ensuring consistency across large-scale datasets.

\subsection{Scoring Model Training}
To build a scoring model that can stably and accurately execute financial scoring rubrics, we fine-tuned Qwen3-32B using SFT and reinforcement training based on RLVR. This section describes the approach from three perspectives: training data construction, training algorithm design, and model performance.
\subsubsection{Training Data Construction}
We rewrote and generalized the above benchmark (including both questions and rubrics) and conducted manual scoring annotation. Annotators were required not only to provide a final score, but also to write a brief “scoring rationale” of approximately one sentence. This practice makes the training data auditable and also lays the foundation for subsequent “reverse chain-of-thought (reverse CoT)” synthesis in the SFT stage \citep{wang2025reverse}.
In manual annotation, score distributions typically exhibit imbalance, with a bias toward mid-to-high score ranges (e.g., 3, 4, and 5). To ensure sufficient samples across all score levels, we introduced the following mechanism: multiple models are used to generate answers to each question, and multiple large models then score these answers according to the rubric using a voting scheme. Samples with high inter-model agreement are selected to augment underrepresented score ranges. This process ensures that the scoring model does not become biased toward a particular score range due to skewed training data, thereby maintaining stable performance in real evaluations.



\subsubsection{Model Training Algorithms}
The scoring model is trained in two stages: (i) explanation learning via supervised fine-tuning (SFT), and (ii) score-alignment optimization via reinforcement learning with verifiable rewards (RLVR).

\paragraph{SFT} 
To equip the model with the ability to identify deficiencies in candidate answers and produce faithful scoring rationales, we adopt a reverse Chain-of-Thought (CoT) synthesis strategy \citep{li2025generation}. 
Specifically, annotators' scoring rationales are treated as reverse reasoning signals to synthesize directed CoT trajectories, enabling the model to jointly predict a score and generate an explicit justification.

\paragraph{RLVR} 
Starting from the SFT checkpoint, we further apply RLVR-based training to improve alignment on scoring tasks. 
We employ an improved variant of DAPO \citep{yu2025dapo} that removes both the KL regularization term and the length penalty, which yields more stable training dynamics.
The reward function is defined along two complementary dimensions:
\begin{itemize}
    \item \textbf{Format reward:} the output must follow a Markdown-compatible JSON schema with complete fields and standardized scores.
    \item \textbf{Accuracy reward:} the predicted score should match the ground-truth human score as closely as possible.
\end{itemize}

\subsubsection{Scoring Model Performance} 
To systematically evaluate the performance of the scoring model, we constructed a test set containing 330 scoring tasks. For each task, the final score is fully consistent across multiple annotators (i.e., high confidence).
We use $score\_diff$ as the primary evaluation metric: a smaller score difference indicates closer alignment with human judgments.

\begin{equation}
\text { ScoreDiff }=\frac{1}{N} \sum_{i=1}^N\left|\hat{s}_i-s_i\right|
\end{equation}

\begin{table}[ht]
\centering
\small 
\caption{Comparison of ScoreDiff across various models on the 330@mean4 test dataset.}
\label{tab:model_comparison}
\begin{tabular}{@{}lcccccccccc@{}}
\toprule
\textbf{Test dataset} & Qwen3 & Gemini-3 & GPT-OSS & Qwen3-235B & GLM & Kimi-K2 & DeepSeek& GPT & Seed-OSS &  \\
\textbf{(330@mean4)} & Max & Pro & 120B & A22B-Think & 4.6 & Think &V3.2  & 5.1& OSS-36B & Ours\\ 
\midrule
ScoreDiff & 0.944 & 0.814 & 0.9062 & 0.824 & 0.947 & 1.042 & 0.929 & 0.867 & 0.887 & \textbf{0.755} \\
\bottomrule
\end{tabular}
\end{table}

\end{document}